%% file: main.tex
\documentclass{article}

\usepackage{arxiv}

\usepackage[utf8]{inputenc} 
\usepackage[T1]{fontenc}    
\usepackage{hyperref}       
\usepackage{url}            
\usepackage{booktabs}       
\usepackage{amsmath}
\usepackage{amsfonts}       
\usepackage{nicefrac}       
\usepackage{microtype}      
\usepackage{lipsum}		    
\usepackage{graphicx}
\usepackage{natbib}
\usepackage{doi}
\usepackage{xcolor}
\usepackage{ulem}
\usepackage{multirow}
\usepackage{xltabular}
\usepackage{threeparttablex}
\usepackage[noabbrev,capitalise,nameinlink]{cleveref}
\usepackage{subfiles}

\AtBeginDocument{}
\AtBeginDocument{}

\DeclareMathOperator*{\argmax}{argmax}

\newcolumntype{S}[1]{>{\raggedright\arraybackslash}p{#1}}
\newcolumntype{L}[1]{>{\raggedright\let\newline\\\arraybackslash\hspace{0pt}}m{#1}}
\newcolumntype{C}[1]{>{\centering\let\newline\\\arraybackslash\hspace{0pt}}m{#1}}

\makeatletter
\newcommand\blueuline{\bgroup\markoverwith
{\textcolor{blue}{\rule[-0.5ex]{2pt}{0.9pt}}}\ULon}
\makeatother

\renewcommand{\arraystretch}{1.5}

\title{InforME: Improving Informativeness of Abstractive Text Summarization With Informative Attention Guided by Named Entity Salience}

\author{
    \href{https://orcid.org/0009-0004-7201-5604}{\includegraphics[scale=0.06]{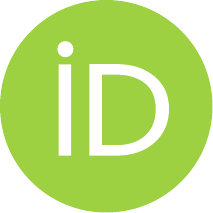}\hspace{1mm}Jianbin Shen} \\
	University of Technology Sydney, Australia \\
	\texttt{Chris.Shen@uts.edu.au} \\
	\And
	\href{https://orcid.org/0000-0001-7179-5208}{\includegraphics[scale=0.06]{orcid.pdf}\hspace{1mm}Christy Jie Liang} \\
	University of Technology Sydney, Australia \\
	\texttt{Jie.Liang@uts.edu.au} \\
	\And
	\href{https://orcid.org/0000-0002-8367-6908}{\includegraphics[scale=0.06]{orcid.pdf}\hspace{1mm}Junyu Xuan} \\
	University of Technology Sydney, Australia \\
	\texttt{Junyu.Xuan@uts.edu.au} \\
}

\date{}

\begin{document}
\maketitle

\begin{abstract}
Abstractive text summarization is integral to the Big Data era, which demands advanced methods to turn voluminous and often long text data into concise but coherent and informative summaries for efficient human consumption.
Despite significant progress, there is still room for improvement in various aspects. One such aspect is to improve informativeness. 
Hence, this paper proposes a novel learning approach consisting of two methods: an optimal transport-based informative attention method to improve learning focal information in reference summaries and an accumulative joint entropy reduction method on named entities to enhance informative salience.
Experiment results show that our approach achieves better ROUGE scores compared to prior work on CNN/Daily Mail while having competitive results on XSum.
Human evaluation of informativeness also demonstrates the better performance of our approach over a strong baseline. 
Further analysis gives insight into the plausible reasons underlying the evaluation results.
Our source code is accessible on GitHub\footnote{\url{https://github.com/74808917/ozizAxRminf6Zyv}.}.
\end{abstract}


\section{Introduction}
Abstractive text summarization (ATS) is a learning task in natural language processing to condense a long document into a summary that is fluent, coherent, relevant, and consistent (e.g., \citealp{kryscinski_neural_2019}). These are commonly accepted criteria for ATS.
Efficient transfer learning enabled by pre-trained language modeling of Transformer (\citealp{vaswani_attention_2017}) based models (e.g., \citealp{devlin_bert_2019}; \citealp{lewis_bart_2020}) has greatly benefited the progress of ATS on these criteria.
Numerous task-specific methods (e.g., \citealp{zhang_attention_2022}; \citealp{gabriel_discourse_2021}), developed to fine-tune pre-trained language models, have further improved and adapted ATS to various data domains, such as news articles and scientific papers.
\par
Among the criteria, relevance concerns generated summaries supported by their source documents.
This is primarily attributable to cross-attention mechanisms devised in encoder-decoder modeling.
For example, in prevalent Transformer-based ATS encoder-decoders, the cross-attention mechanisms intuitively assign each decoded token's latent state with a distribution over the source document sequence token latent states.
This teaches models to generate summaries correlated to their source documents.
Although it is essential, this form of learning relevance may not always be optimal for informativeness from a reference summary viewpoint because it could ignore focal information in reference summaries that are informative but less statistically correlated with the source documents otherwise, for example, extrinsic knowledge.
This inadequacy is evident in generated summaries, which often contain less informative content, and it is not uncommon to miss out on substantial information.
For example, given the following elliptically shortened long source document,
\begin{quote}
\textit{``... But it is not impossible that our decision to leave the European Union could end up being judged in the European Court in Luxembourg. ...could, in theory, fight to keep us in their clutches. ...''},
\end{quote}
and its reference summary,
\begin{quote}
\textit{``Irony of ironies, is it possible that the European Court could block us from leaving the European Union?''},
\end{quote}
A pre-trained model may generate a summary on the same source document as follows,
\begin{quote}
\textit{``If you are an ardent Brexiteer, stop before you crack open the champagne.''}.
\end{quote}
Clearly, the model-generated summary is much less informative compared to the reference summary.
\par
Recognizing the gap, researchers have proposed solutions to improve the informativeness of ATS (e.g., \citealp{wang_friendly_2020}; \citealp{fu_document_2020}).
The approaches typically integrate topic modeling, which has long been used in extractive text summarization (e.g., \citealp{shenouda_summvd_2022}; \citealp{wu_topic_2017}). 
Many of them adopt statistical topic extraction methods. The most used statistical method is Latent Dirichlet Allocation (e.g., \citealp{narayan_dont_2018}; \citealp{huang_legal_2020}), whereas 
Poisson Factor Analysis (e.g., \citealp{wang_friendly_2020}) is also used to learn topic words. 
The neural topic models (e.g., \citealp{fu_document_2020}; \citealp{ma_t-bertsum_2022}) are recent alternatives to statistical methods for topic-relevant ATS modeling.
However, topic modeling commonly imposes a fixed number of topics. This presumption makes topic modeling-based ATS sensitive to the number of topics and affects the applicability of the learned ATS models.
\par
This paper proposes a novel learning approach that includes an optimal transport-based informative attention method with an accumulative joint entropy reduction method on named entities.
Our approach has the following advantages over topic modeling-based prior work.
We integrate our methods into ATS modeling in end-to-end training seamlessly, whereas ATS modeling integrated with topic modeling is a two-stage process.
Our approach is free of the aforementioned limitations imposed by topic modeling in ATS modeling.
Our learning approach reconciles the reference summary perspective with the source document relevance in training, while prior work focuses on topics supported by the source documents.
In summary, our approach has the following contributions,
\begin{enumerate}
    \item We propose an optimal transport (OT)-based informative attention method. It is devised as a reverse cross-attention complementary to the cross-attention of the existing encoder-decoder modeling.
    It aims to learn focal information in reference summaries that could otherwise be ignored.
    \item We further propose an accumulative joint entropy reduction (AJER) method.
    It is based on information theory, consisting of an accumulative negative information gain component and a joint entropy reduction component.
    Utilizing named entities, the method enhances the salience of named entities in the model's latent space, which in turn guides the proposed informative attention method efficiently.
    \item Our experiments demonstrate that our approach outperforms prior work on summary ROUGE scores with CNN/Daily Mail while maintaining competitive results on XSum.
    Human evaluation of informativeness also shows that our approach achieves better results than a strong baseline.
    The further analysis sheds light on the plausible reasons behind the evaluation results.
\end{enumerate}

\section{Related Work}
To improve the informativeness of ATS, researchers typically incorporate topic modeling into ATS modeling.
Statistical topic modeling, for example, using Latent Dirichlet Allocation, has been widely used with various ATS models, such as LSTM-based models (e.g., \citealp{huang_legal_2020}), convolutional models (e.g., \citealp{khanam_joint_2021}),
and more recent Transformer-based models (e.g., \citealp{wang_friendly_2020}).
The different ATS models open novel ways to incorporate extracted topics into ATS modeling. \citet{narayan_dont_2018} pass the topic distributions directly to the model as additional inputs, whereas \citet{pan_sequence--sequence_2019} infuse the model's attention mechanism with the topic distributions.
\citet{ailem_topic_2019} formulate the topic distributions into pointer-generator-based conditional probability estimation.
\citet{huang_legal_2020} further introduce a topic context learned from topic distributions into their pointer-generator.
Utilizing ConceptNet, \citet{khanam_joint_2021} integrate two extra knowledge-infused attention channels: input-topic knowledge attention and topic knowledge-summary attention.
In comparison, \citet{wang_friendly_2020} extend the Transformer's cross-attention with a topic semantic attention head and leverage the topic proportional factor from Poisson factor analysis to modulate the output of the integrated cross-attention.
\par
Neural topic modeling (NTM) is an alternative to statistical topic modeling to learn topic distributions.
Neural topic models are mostly based on the variational autoencoders (VAE).
The inputs to NTM are usually either input embeddings or the encoder's output representations.
Likewise, different ATS models open researchers to various approaches of integrating learned topics into topic-aware decoding.
For example, \citet{fu_document_2020} compute a topic context vector from topic distributions and add it to the decoder's linear transformation to acquire the decoding classification logits, whereas \citet{zheng_topic-guided_2021} extend the decoder to combine the decoding classification logits head with a devised topic model head argument on the topic distributions.
On the other hand, \citet{ma_t-bertsum_2022} develop a two-stage encoder-decoder 
using the topic-infused embeddings.
The first stage encodes the topic-aware sentence representations by binary classification. 
The sentence representations then gate the second stage encoder's outputs for decoding.
Following the same idea of topic-aware sentences, \citet{jiang_gatsum_2022} learn topic-aware sentence representations as inputs to the cross-attention, using a topic-sentence bipartite graph of the topic representations and the sentence representations.
\par
Differing from VAE-based topic models, \citet{aralikatte_focus_2021} develop embedding-transformed topic distributions, which are combined to form a dynamic vocabulary bias vector. The bias vector is then formulated into the generative token probability distribution estimation.
The authors practically fix the bias components to a predefined number of sampled tokens.
Some researchers have also explored language models for topic extraction.
For instance, \citet{zheng_controllable_2020} leverage a pre-trained BERT-NER to extract named entities in source documents as topics, but their BiLSTM decoder is conditioning on a fixed number of selected entities constrained by their decoder modeling.

\section{Background}
Transformer-based encoder-decoders have an autoregressive formulation as a conditional probability estimation problem:
\begin{equation}
\label{eq:conditional_generation_probability}
\begin{aligned}
    p(\hat{y}_t|\theta) = p_{\theta}(\hat{y}_t|y_{1:t-1},\mathbf{x})
\end{aligned}
\end{equation}
where a parameterized model $\theta$ generates a probability distribution\footnote{We may omit the conditional notation for simplicity without loss of generality when there is no ambiguity.} $p_{\theta}(\hat{y}_t)$ at step $t$ conditional on the decoding inputs $y_{1:t-1}$, a teacher-forcing learning strategy (\citealp{williams_learning_1989}), and the source document sequence $\mathbf{x}$.
The learning objective becomes to maximize the generative likelihood:
\begin{equation}
\label{eq:optimize_params_to_maximize_p}
\begin{aligned}
    \hat{y}_t =
    \argmax_{\theta}(p_{\theta}(\hat{y}_t|y_{1:t-1},\mathbf{x})).
\end{aligned}
\end{equation}
This maximization problem is translated into a negative log-likelihood minimization problem using the cross-entropy loss function for training:
\begin{equation}
\label{eq:mle_learning_objective}
\begin{aligned}
    \mathcal{L}_{\mathrm{MLE}} 
    &= -\mathbb{E}_{\hat{y}_{t} \sim p_{\theta}}[y_t\log \hat{y}_t]
    \approx 
    -\frac{1}{n} \sum_{t=1}^{n} y_t \log p_{\theta}(\hat{y}_t)
\end{aligned}
\end{equation}
where $y_t$ is the ground truth at step $t$, and $n$ is the length of the generated sequence.

\begin{figure}[t!]
\centering
  \includegraphics[width=0.7\columnwidth]{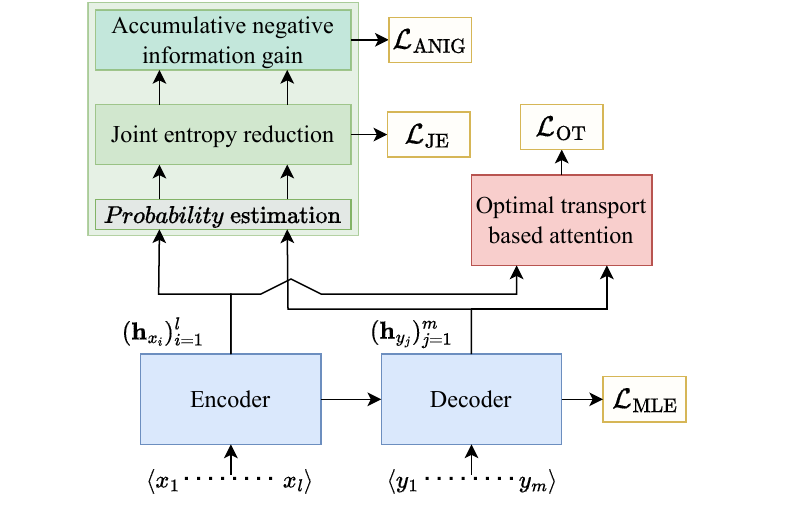}
  \caption{Illustration of an encoder-decoder with our methods, including optimal transport-based informative attention (carmine block) and accumulative joint entropy reduction (tealish block).}
  \label{fig:ats_informe}
\end{figure}

\section{Our Methods}
\autoref{fig:ats_informe} illustrates our learning approach, which extends an encoder-decoder with our optimal transport-based informative attention and accumulative joint entropy reduction methods.

\subsection{Optimal Transport-Based Informative Attention}
Learning summary generation should not only be relevant to source documents but also be informative concerning the reference summaries.
This prompts the idea of reverse cross-attention.
To this end, we devise an informativeness regularization as the optimal transport (OT) problem using Kantorovich's formulation (e.g., \citealp{kolouri_optimal_2017}), which aims to find an optimal transport plan. 
The metric for the transport plan is Wasserstein distance (or Kantorovich distance).
In our context, it is about moving semantic information in such a way that the source document content only `interested' by the summary distribution is retained for summary generation.
For an $l$-length source document and the corresponding $m$-length summary, the discrete optimal transport problem, as a minimizer, can be expressed as follows:
\begin{equation}
\begin{aligned}
\label{eq:ot_plan_minimizer}
    K(x,y) =& 
    \min_{\pi}\sum_{i=1}^{l} \sum_{j=1}^{m} \pi(x_i,y_j) c(x_i,y_j),
    \\
    \text{s.t.} 
    \:
    \sum_{j=1}^{m} & \pi(x_i,y_j) = \pi(x_i), 
    \\
    \sum_{i=1}^{l} & \pi(x_i,y_j) = \pi(y_j),
    \\
    & \pi(x_i,y_j) \ge 0
\end{aligned}
\end{equation}
where $\pi(\cdot,\cdot)$ is joint distribution as a coupling matrix, and $c(\cdot,\cdot)$ is cost function.
The joint distributions achieving the infimum are called optimal transport plans.
\par
To realize an optimal transport, we first formulate the cost function as follows.
Given the latent states (or representations) of the source document $\mathbf{h}_{x}$ and the corresponding summary $\mathbf{h}_{y}$, the cost function is defined as an L2 norm:
\begin{align}
\label{eq:ot_cost}
    c_{(x_i, y_j)} = ||\mathbf{h}_{x_i} - \mathbf{h}_{y_j}||_2.
\end{align}
For each source token $\mathbf{h}_{x_i}$, its distances to the summary tokens $\{\mathbf{h}_{y_j}\}_{j=1}^{m}$ are computed.
\par
Formulating the joint distribution (the plan) using a coupling method is crucial to learning an optimal transport that maximizes information concerning reference summaries.
The orthogonal property of bi-linearity means that a source token representation similar to the summary token representation(s) would become salient, whereas it would be suppressed.
This provides a simple yet effective coupling method for learning the focal information from source documents relevant to the summaries. 
So, we formulate the joint distribution based on the bilinear transformation of $(x,y)$:
\begin{align}
    \mathrm{T}_{i,j} &= \mathbf{h}_{x_i} W_{i,j} \mathbf{h}_{y_j}
\end{align}
where $W_{i,j} \in W$ is the weight of the coupling pair.
The joint distribution is then defined as:
\begin{align}
\label{eq:ot_joint_distribution}
    \pi_{(x_{i}, y_{j})} &= \frac{e^{z_{i,j}}}{\sum_{j^{\prime}} e^{z_{i,j^{\prime}}}}, \; z_{i,j} \in \mathrm{T}_{i}
\end{align}
where a probability distribution over the summary token representations is computed for each source document token representation.
Substituting the terms in \autoref{eq:ot_plan_minimizer} with \autoref{eq:ot_cost} and \autoref{eq:ot_joint_distribution}, we derive the informative attention learning objective as an average of the transport plan minimizer:
\begin{equation}
\label{eq:ot_learning_objective}
    \mathcal{L}_{\mathrm{OT}} = \frac{1}{l \times m} K(x,y).
\end{equation}

\subsection{Accumulative Joint Entropy Reduction}
\label{subs:structural_intergrity_of_named_entities}
The method gives rise to informative attention learning through named entity latent salience.
Named entities\footnote{We follow the classification of named entity recognition.} provide the focal points since document content is commonly laid out through them.
Increasing the salience of named entities implies reducing their uncertainty from an information theory viewpoint.
\par
A named entity often consists of multiple words. Furthermore, token encoding schemes like byte-pair encoding (\citealp{sennrich_neural_2016}) employed in ATS modeling may segment a word into several tokens.
This means that our method should reduce the uncertainty of the named entity tokens collectively as well as the uncertainty in their sequential structure.
Our method addresses both uncertainties holistically, as detailed as follows.
\par
Uncertainty is measured by (Shannon) entropy in information theory.
The higher entropy, the higher uncertainty\footnote{We use entropy and uncertainty interchangeably when there is no ambiguity.}.
The information theory (e.g., \citealp{polyanskiy_information_2022}) also 
states that conditioning reduces entropy (or uncertainty). Furthermore, under any stationary process $X=(X_1,...X_n)$, it satisfies:
\begin{equation}
    H(x_n|X_{1:n-1}) \le H(x_{n-1}|X_{1:n-2})
\end{equation}
Named entities follow a stationary process and are conditional.
This suggests that each conditional named entity token prediction should be no more uncertain than that at the previous step.
The empirical study by \citet{xu_understanding_2020} also supports the theory in the context of ATS.
This permits us to formulate uncertainty reduction for named entities as follows.
\par
Let us express a named entity token sequence as $x_{1:n}$, and assume the sequence is a Markov chain for notation simplicity, the conditional entropy of a named entity token can be expressed as\footnote{Entropy can then be defined for any Markov processes.}:
\begin{equation}
\label{eq:conditional_entropy_of_named_entity_token}
\begin{cases}
    H(x_i|x_{i-1}), & i > 1,
    \\
    H(x_i), & i = 1.
\end{cases}
\end{equation}
To simplify our discussion without loss of generality,
we introduce notations $H(x_{0}) = 0$ and $H(x_1|x_{0}) = H(x_1)$.
We can thus express a conditional uncertainty inequality as:
\begin{equation}
\label{eq:conditioning_reduce_uncertainty_expression}
    H(x_{i+1}|x_{i}) \le H(x_{i}|x_{i-1}), \; i \ge 1.
\end{equation}
After rearranging, we obtain an adjacent step entropy difference as follows:
\begin{equation}
\label{eq:adjacent_information_gain}
\begin{aligned}
    \triangle H(x_{i}, x_{i+1}) = 
    H(x_{i+1}|x_{i})
     - 
    H(x_{i}|x_{i-1})
    .
\end{aligned}
\end{equation}
We reframe this entropy reduction as an adjacent information gain.
Meanwhile, following the chain rule for entropy, the joint entropy of the sequence can be defined as:
\begin{equation}
\label{eq:joint_entropy_of_sequence}
    H(x_{1:n}) = \sum_{i=1}^n H(x_i|x_{i-1}).
\end{equation}
\par
From \autoref{eq:adjacent_information_gain} and \autoref{eq:joint_entropy_of_sequence}, we derive a joint information gain:
\begin{equation}
\label{eq:joint_information_gain}
\begin{aligned}
    \triangle H(x_{2:n}) = & \sum_{i=2}^{n} 
    (H(x_i|x_{i-1}) - H(x_{i-1}|x_{i-2})).
\end{aligned}
\end{equation}
Maximizing this quantity reduces the relative uncertainty in the conditional prediction of the generative sequence.
This maximization quantity is translated into a negative joint information gain for minimization training:
\begin{equation}
\label{eq:negative_joint_information_gain}
\begin{aligned}
    \mathcal{L}_{\text{NIG}} = -\triangle H(x_{2:n}).
\end{aligned}
\end{equation}
Drawing on the conditioning entropy theory, we expect the relative uncertainty to be higher at the beginning of an entity and lower toward the end of the entity sequence. It would be reasonable to apply hyperparameters to weight each term in the summation to regularize the uncertainty reduction for efficient learning. However, searching for proper hyperparameters could be time-consuming. Instead, we extend \autoref{eq:negative_joint_information_gain} to an accumulative formulation, named accumulative negative information gain:
\begin{equation}
\label{eq:nig_regularizer}
\begin{aligned}
    \mathcal{L}_{\mathrm{ANIG}} &= - \frac{1}{n} \sum_{i=2}^{n} \sum_{j=2}^{i} \triangle H(x_{2:j}).
\end{aligned}
\end{equation}
This has the effect of penalizing more in the beginning and then weighting the penalty less gradually.
\autoref{eq:nig_regularizer} encourages the learning model to conditionally choose the next token that is less uncertain than the current token. It thus reduces the uncertainty in the token sequential structure.
However, it may not guarantee efficiency in reducing the absolute uncertainty of choosing a named entity. This can be amended by using the joint entropy as a regularization term too, defined as:
\begin{equation}
\label{eq:joint_entropy_regularizer}
    \mathcal{L}_{\mathrm{JE}} = \frac{1}{n} \sum_{i=1}^n H(x_i|x_{i-1}).
\end{equation}
\autoref{eq:nig_regularizer} and \autoref{eq:joint_entropy_regularizer} form two components of the accumulative joint entropy reduction (AJER).
The double-dose formulation safeguards the integrity of named entity estimations.
The AJER can be applied to named entities in both source documents and summaries due to the conditionality inherent in the entities.
The autoregressive nature of the decoder further makes the conditionality model-explicit on summary entities.

\subsection{Total Learning Objective}
Composing of \autoref{eq:mle_learning_objective}, \autoref{eq:ot_learning_objective}, \autoref{eq:nig_regularizer}, and \autoref{eq:joint_entropy_regularizer}, we arrive at the total learning objective:
\begin{equation}
\label{eq:total_learning_objective}
\mathcal{L} = \mathcal{L}_{\mathrm{MLE}} + 
\alpha_{\mathrm{OT}}\mathcal{L}_{\mathrm{OT}} + 
\alpha_{\mathrm{ANIG}}\mathcal{L}_{\mathrm{ANIG}} + 
\alpha_{\mathrm{JE}}\mathcal{L}_{\mathrm{JE}}
\end{equation}
where $\alpha_{\mathrm{OT}}$, $\alpha_{\mathrm{ANIG}}$, and $\alpha_{\mathrm{JE}}$ are hyperparameters to control objective influences.

\section{Experiment Results and Analysis}
\begin{table*}[t!]
\centering
\begin{threeparttable}
\begin{tabularx}{\textwidth}{|>{\raggedright\arraybackslash}p{2.5in}|X X X|X X X|}
 \hline
 \multirow{2}{*}{\textbf{Model}} 
 & 
 \multicolumn{3}{c|}{\textbf{CNNDM Test Set}\tnote{}} 
 &
 \multicolumn{3}{c|}{\textbf{XSum Test Set}\tnote{}} 
 \\
 \multicolumn{1}{|c|}{} 
 & R-1 & R-2 & R-L\tnote{1}
 & R-1 & R-2 & R-L\tnote{1}
 \\
 \hline
 TAG+Cov (\citealp{ailem_topic_2019})
 & 40.06 & 17.89 & 36.52
 & N/A & N/A & N/A
 \\
 \hline
 BertSUM+TA (\citealp{wang_friendly_2020})
 & 43.06 & 20.58 & 39.67
 & 39.77 & 17.39 & 32.39
 \\
 \hline
 TAS (\citealp{zheng_topic-guided_2021})
 & 44.38 & 21.19 & 41.33
 & 44.63 & 21.62 & 36.77
 \\
 \hline
 KTOPAS (\citealp{khanam_joint_2021})
 & 42.10 & 20.01 & 38.45
 & N/A & N/A & N/A
 \\
 \hline
 PEGFAME (\citealp{aralikatte_focus_2021})
 & 42.95 & 20.79 & 39.90
 & 45.31 & 22.75 & 37.46
 \\
 \hline
 T-BERTSUM(ExtAbs/Abs) (\citealp{ma_t-bertsum_2022})\tnote{2}
 & 43.06 & 19.76 & 39.43
 & 39.90 & 17.48 & 32.18
 \\
 \hline
 GATSum (\citealp{jiang_gatsum_2022})
 & 44.46 & 21.32 & 39.84
 & 44.60 & 21.53 & 36.66
 \\
 \hline
 BART-large (\citealp{lewis_bart_2020})\tnote{}
 & 44.16 & 21.28 & 40.90 
 & 45.14 & 22.27 & 37.25
 \\
 \hline
 \hline
 BART-large/OT\tnote{}
 & 44.67 & 21.16 & 41.59 
 & 45.16 & 21.69 & 36.54 
 \\
 \hline
 BART-large/AJER\tnote{}
 & 44.67 & 21.31 & 41.58
 & 44.83 & 21.51 & 36.17
 \\
 \hline
 BART-large*/OT+AJER\tnote{}
 & \textbf{44.75} & \textbf{21.54} & \textbf{41.69} 
 & 45.08 & 21.58 & 36.22 
 \\
 \hline
\end{tabularx}
\caption{ROUGE evaluation (on CNNDM and XSum).
CNNDM: 11490 samples. XSum: 11334 samples.
1. Our summary-level R-L scores (equivalent to the R-Lsum ROUGE metric this paper uses).
2. T-BERTSUM(ExtAbs) for CNNDM, and T-BERTSUM(Abs) for XSum.
\label{tab:rouge_eval_with_cnndm_xsum}}
\end{threeparttable}
\end{table*}

\subsection{Dataset}
\label{subsect:dataset}
We choose two typical English benchmark datasets commonly used in ATS research, CNN/Daily Mail (CNNDM) (\citealp{see_get_2017}; \citealp{hermann_teaching_2015}) and XSum (\citealp{narayan_dont_2018}).
CNNDM's reference summaries tend to be extractive and intrinsic, while those of XSum are extremely abstractive and extrinsic, as the extrinsic characteristic is also studied and quantified by \citet{lu_multi-xscience_2020}.
We use Stanford CoreNLP (v4.4.0) named entity recognition (NER) parsing tool\footnote{\url{https://stanfordnlp.github.io/CoreNLP}.} to acquire named entities from the training datasets. The data preparation is detailed in \Cref{app:sect:data_preparation}.

\subsection{Implementation}
We use BART (\citealp{lewis_bart_2020}) as our backbone encoder-decoder. Our implementation adopts Hugging Face's BART implementation\footnote{\url{https://huggingface.co/transformers/v4.9.2/model\_doc/bart.html}.} with the pre-trained BART-large weight profiles\footnote{\url{https://huggingface.co/facebook/bart-large-cnn} for CNNDM, and \url{https://huggingface.co/facebook/bart-large-xsum} for XSum.}.
Note that the model evaluation is conducted on the trained backbones solely, as our methods are only used to train the backbone models.
The pre-trained BART-large models are also used as baselines for evaluation comparison.
The key implementation details are given in \Cref{app:sect:key_implementation}.

\subsection{Results and Analysis}
We conduct ROUGE evaluation, factuality consistency evaluation, and human evaluation.
Our analysis covers them to gain insights into the results and plausible reasons.

\subsubsection{ROUGE Evaluation}
We first present ROUGE\footnote{The implementation of ROUGE metrics (\citealp{lin_rouge_2004}) used in this paper is from Hugging Face's Python package datasets.} results in \autoref{tab:rouge_eval_with_cnndm_xsum}. The table includes prior work for comparison, starting with several recent topic-aware ATS research outcomes that have achieved high ROUGE scores, followed by the published BART-large results for ATS.
Separated by double lines are our experiments, which include the ablation study results of OT and AJER.
Our OT+AJER-trained backbone achieves the best results compared to prior work on CNNDM, while our OT+AJER-trained backbone on XSum also has competitive results.
We think that the characteristics of XSum may impact the performance of our methods.
Later, our human evaluation explores the impact and draws on plausible reasons behind the results.

\paragraph{Ablation Study}
The ablation study trains our backbone models with OT and AJER, respectively. 
Comparing the ablation results shown in \autoref{tab:rouge_eval_with_cnndm_xsum}, we see that the OT+AJER-trained model achieves better results on CNNDM, while the OT-trained model edges ahead on XSum.
We will also explore the reasons underlying the observation here in our human evaluation.

\subsubsection{Automatic Factuality Consistency Evaluation}
\label{sect:auto_fact_consistency_evaluation}
\begin{table*}[t!]
\centering
\normalsize
\begin{threeparttable}
\begin{tabularx}{0.8\columnwidth}
{|>{\raggedright\arraybackslash}p{1.5in}|X|X|}
 \cline{1-3}
 \multirow{2}{*}{\textbf{Model}\tnote{}}
 & \multicolumn{2}{c|}{\textbf{QuestEval}}
 \\
 \cline{2-3}
 & \multicolumn{1}{c|}{CNNDM $\mathbf{\mu} (\%)$}
 & \multicolumn{1}{c|}{XSum $\mathbf{\mu} (\%)$}
 \\
 \cline{1-3}
 BART-large\tnote{}
 & 46.5
 & 45.1
 \\
 \cline{1-3}
 BART-large* /OT+AJER\tnote{}
 & \textbf{47.0}
 & 44.7
 \\
 \cline{1-3}
\end{tabularx}
\caption{QuestEval mean score statistic over the summaries generated from the CNNDM and XSum test sets (measured against reference summaries).
\label{tab:questeval_evaluation_on_cnndm_xsum_vs_ref}
}
\end{threeparttable}
\end{table*}
\begin{table*}[t!]
\centering
\setlength{\tabcolsep}{3pt}
\begin{threeparttable}
\begin{tabularx}{0.8\textwidth}
{|>{\raggedright\arraybackslash}p{1.5in}|X|X|X|X|}
 \cline{1-5}
 \multirow{3}{*}{\textbf{Model}\tnote{}}
 & \multicolumn{2}{c|}{\textbf{CNNDM} (60 samples)}
 & \multicolumn{2}{c|}{\textbf{XSum} (60 samples)}
 \\
 \cline{2-5}
 & \multicolumn{1}{c|}{No.}
 & \multicolumn{1}{c|}{\%}
 & \multicolumn{1}{c|}{No.}
 & \multicolumn{1}{c|}{\%}
 \\
 \cline{1-5}
 BART-large\tnote{}
 & 11/60
 & 18.33
 & 9/60
 & 15.00
 \\
 \cline{1-5}
 BART-large*/OT+AJER\tnote{}
 & \textbf{18}/60
 & \textbf{30.00}
 & \textbf{19}/60
 & \textbf{31.66}
 \\
 \cline{1-5}
\end{tabularx}
\caption{Human evaluation of better informativeness on 60 randomly drawn samples of generated summaries from CNNDM and XSum, respectively.
\label{tab:human_evaluation_informativeness_cnndm_xsum}}
\end{threeparttable}
\end{table*}
Factuality consistency metrics are possibly the most relevant choice for measuring informativeness semantically.
Several automatic factuality consistency metrics have been developed over the years (e.g., \citealp{kryscinski_evaluating_2020}; \citealp{durmus_feqa_2020}; \citealp{scialom_questeval_2021}).
Given that our methods reconcile the reference summaries, question-answering model-based metrics are proper for the purpose. For that reason, we choose QuestEval (\citealp{scialom_questeval_2021}) and evaluate model-generated summaries against the golden references.
We compare our OT+AJER-trained backbone with the BART-large baseline, given that the pre-trained baseline implementation is available and utilized as our backbone model.
The mean scores are shown in \autoref{tab:questeval_evaluation_on_cnndm_xsum_vs_ref}.
The results show that the OT+AJER-trained model performs better than the baseline on CNNDM, while the baseline edges ahead on XSum. The results are in agreement with the ROUGE evaluation in terms of performance.

\subsubsection{Human Evaluation}
To further ascertain automatic evaluations and gain a better understanding of plausible reasons underlying the results, we conduct our human evaluations on both informativeness and factuality.

\paragraph{Informativeness Evaluation}
This evaluation is challenging in that model-generated summaries may contain extrinsic information not evidenced in the articles and reference summaries. They may also contain factual errors. Some errors may obscure informativeness more than others.
It is also challenging because reference summaries could occasionally be uninformative or erroneous.
So, we set out a few key rules to ensure the evaluation is more objective, including 
taking into account factual error impact on the informative explication, and 
consolidating reference summaries with the information the articles convey instead of relying on reference summaries solely. 
The evaluation does not take into consideration the extrinsic information in model-generated summaries, which is not evident in source documents and reference summaries.
They will be assessed in a human evaluation of factuality when extrinsic issues are investigated.
The evaluation guidelines are detailed in \Cref{app:subsect:definition_of_criteria}.
Note that we `normalize' whitespaces in model-generated summaries and detail reasons in \Cref{post_formatting_summaries_for_informativeness_evaluation}.
\par
To facilitate the evaluation, we have developed a user interface-based evaluation tool and procedures as detailed in \Cref{app:subsect:human_evaluation_user_interface}.
\par
The results are shown in 
\autoref{tab:human_evaluation_informativeness_cnndm_xsum}. The OT+AJER-trained backbones generate more informative summaries than the baselines on both CNNDM (by \texttt{\char`\~}$11\%$) and XSum (by \texttt{\char`\~}$16\%$), respectively.
The resulting disparities between the OT+AJER-trained backbones and the pre-trained baselines could be a good indication of sampled token distributional shifts.

\paragraph{Factuality Evaluation}
\begin{table*}[t!]
\centering
\begin{threeparttable}
\begin{tabularx}{0.8\columnwidth}
{|>{\raggedright\arraybackslash}p{1.3in}|X|X|X|X|}
 \hline
 \multirow{2}{*}{\textbf{Error Type}}
 & \multicolumn{2}{c|}{\textbf{CNNDM}}
 & \multicolumn{2}{c|}{\textbf{XSum}}
 \\
 \cline{2-5}
 & Baseline\tnote{2}
 & Ours\tnote{3}
 & Baseline\tnote{2}
 & Ours\tnote{3}
 \\
 \hline 
 Entity extrinsic\tnote{1}
 & 1
 & 1
 & 2
 & 8 (5+2+1)\tnote{4}
 \\
 \hline
 \hline 
 Name
 & 2
 & 1
 & 6
 & 5
 \\
 \hline
 \multirow{1}{*}{Event}
 & 0
 & 0
 & 1 
 & 1
 \\
 \hline
 \multirow{1}{*}{Event-time}
 & 0
 & 0
 & 1 
 & 1
 \\
 \hline
 \multirow{1}{*}{Location}
 & 0
 & 1
 & 3 
 & 2
 \\
 \hline
 \multirow{1}{*}{Number}
 & 0
 & 0
 & 10
 & 8
 \\
 \hline 
 \multirow{1}{*}{Other}
 & 3
 & 3
 & 11 
 & 5
 \\
 \hline
 \hline
\multirow{1}{*}{\textit{\textbf{Total}}}
 & \textit{6}
 & \textit{6}
 & \textit{34}
 & \textit{\textbf{30 (25)}}
 \\
 \hline
 \end{tabularx}
\caption{Human evaluation of factuality on the same 60 randomly sampled generated summaries (CNNDM and XSum, respectively).
1. The entity extrinsic covers person names, events, and locations.
2. BART-large baseline.
3. BART-large*/OT+AJER.
4. Five of them are factual, two are erroneous, and one is inconclusive.
}
\label{tab:human_eval_factuality_on_cnndm_xsum}
\end{threeparttable}
\end{table*}
As we expect that our methods could alter the generative distributions considerably, we are curious about what it actually means for ATS factuality.
We thus conduct the human evaluation on factuality to answer the question and further obtain some insights underlying the answers.
The evaluation is a non-trivial task because factual issues or hallucinations can have various forms (e.g., \citealp{maynez_faithfulness_2020}; \citealp{tang_understanding_2023}). \citet{shen_mitigating_2023} have demonstrated that their human evaluation of factuality using syntactic agreement categorization gives a better understanding of hallucination forms and plausible underlying causes. We follow their approach by tailoring it to our NER-related focus.
\par
The results are shown in \autoref{tab:human_eval_factuality_on_cnndm_xsum}.
It is noticeable that our OT+AJER-trained backbone incurs a considerable number of extrinsic entity issues on XSum that are not seen in either corresponding source documents or reference summaries.
As extrinsic entities can be factual (e.g., \citealp{cao_hallucinated_2022}), we are interested in knowing how many of these extrinsic issues are factual.
After investigating the cases using external resources (e.g., Google search), it turns out that five of them are factual and informative, two are erroneous, and one is inconclusive. Taking into account the five factual extrinsic entities, we see the overall factual error counts reduced to nine fewer errors than the baseline on XSum.
\par
The finding suggests that our methods may be capable of extrinsic data mining within a dataset, whereas extrinsic data mining typically employs external knowledge bases.
We hypothesize that the salience boosted by the AJER method helps the OT method to effectively `build' the correlational linkage between entities in source documents and extrinsic entities in reference summaries in a global context, both document-wide and corpus-wide.
The hypothesis would be consolidated if the summaries generated by the OT-trained model had fewer factual extrinsic entities.
To this end, we analyze the extrinsic entities in the summaries generated by the OT-trained model on the same numbered random samples.
We find five extrinsic entity issues, among which two are factual and three are erroneous.
This result supports our hypothesis.
These findings might partially explain the results on XSum in \autoref{tab:rouge_eval_with_cnndm_xsum} where the scores from the OT+AJER-trained model are lower than those of the OT-trained model and some prior works. They could similarly reason the results on XSum in \autoref{tab:questeval_evaluation_on_cnndm_xsum_vs_ref}.
We believe that the model's increased ability to utilize globally learned extrinsic knowledge in composing summaries gives rise to a capacity closely analogous to human intelligence in writing summaries using extrinsic knowledge.
We provide the five extrinsic but factual entity-related summaries in \Cref{app:subsect:extrinsic_factual_examples}.
\par
Additionally, the summaries generated by our trained model incur fewer categorized syntactic agreement errors on XSum than those by the baseline, including person name, location, number, and `Other' categories. 
We believe this error reduction across categories is also attributable to the integrated AJER method, which enhances the salience of named entities in their contexts within the model latent space and thus reduces the uncertainty in their conditional probability estimations.
\par
Meanwhile, \autoref{tab:human_eval_factuality_on_cnndm_xsum} shows that the OT+AJER-trained model also fends off hallucinations well on CNNDM, on a par with the baseline.
The rationale for the analysis on XSum above is also applicable to the results on CNNDM.
The difference is that the entities in both source documents and reference summaries of CNNDM are largely from the same distribution.
Thus, incurring much less extrinsic information may explain the better scores resulting from the OT+AJER-trained model on CNNDM in \autoref{tab:rouge_eval_with_cnndm_xsum} and \autoref{tab:questeval_evaluation_on_cnndm_xsum_vs_ref} because of more gram overlapping lexically and similar semantically in the model latent space between the OT+AJER-trained model-generated summaries and the reference summaries, respectively.
\par
We also notice that the models trained on CNNDM have produced much fewer errors than the models trained on XSum.
This suggests that existing ATS modeling is efficient at learning from extractive and intrinsic datasets but weak on extremely abstractive and extrinsic datasets.
Our approach may have provided a data-efficient way to reduce the gap, given that the OT+AJER-trained model reduces the number of errors on XSum compared to the baseline.

\section{Conclusion}
This paper proposes an optimal transport-based informative attention method guided by an accumulative joint entropy reduction method, aiming at improving the informativeness of ATS.
Our experiments and analysis have demonstrated that our methods can improve ATS on informativeness and factuality, and are applicable to two typical but very different characteristic datasets.

\section*{Limitations}
Our research in this paper is confined to English news article documents, even though the chosen datasets represent typical characteristics of ATS.
Our future investigation may expand our methods to other document types.
The bias exposure will also be closely examined in our future work through the lens of empirical experiments with different learning objective weight configurations and the study of mathematical bounds.
The evaluation methods, including auto-metric evaluations and human assessment, have their strengths and weaknesses.
We hope that the combination of the auto-metric evaluations and human evaluations on informativeness and factuality may have improved our overall analysis.
A study with larger backbone models may also be undertaken in the future when adequate computational resources become available.

\section*{Ethics Statement}
To the best of our knowledge, we have attributed our work to the previous research and implementations on which this paper depends, either in the main script or in the appendices. 
Any changes to third-party implementations follow the license permits of those implementations.
One author conducts the human evaluations.
Our work and experiment results are all genuine.

\bibliographystyle{unsrtnat}
\bibliography{references}

\appendix
\subfile{appendix}

\end{document}

%% file: appendix.tex
\section{Experiment Setup}

\subsection{Data Preparation}
\label{app:sect:data_preparation}
We download both CNNDM and XSum datasets as JSON files (train, validation, and test) using Hugging Face's Python package datasets.
We utilize Stanford CoreNLP's named entity recognition tool to obtain named entity annotation data, including entity names and their word positions in sentences and documents.
The annotated NER entities used for this research include TITLE, PERSON, ORGANIZATION, NATIONALITY, RELIGION, IDEOLOGY, DEGREE, DATE, TIME, DURATION, LOCATION, CITY, STATE\_OR\_PROVINCE, COUNTRY, NUMBER, MONEY, PERCENT, ORDINAL, CAUSE\_OF\_DEATH, CRIMINAL\_CHARGE.
To minimize potential erroneous annotations, we only include entities whose length is not greater than ten.
We thereafter transform them into our training datasets of both CNNDM and XSum.
\par
The Stanford CoreNLP uses Penn Treebank (\citealp{marcus_building_1993}), which segments words, to produce the parsing tree for NER annotation.
Therefore, our training source documents and summaries are built from the annotated word sequences.
We further preprocess the token encodings\footnote{\url{https://huggingface.co/transformers/v4.9.2/model_doc/bart.html\#transformers.BartTokenizer}.}
of the annotated training documents and summaries for runtime efficiency.
\par
To map the word-level annotated named entities to the token-level encodings during fine-tuning (training), a word-token map (graph) is created for each annotated source document and summary. 
\par
The token encodings of validation and test datasets (CNNDM and XSum) are also prepared without annotation.
\autoref{app:tab:dataset_size_cnndm_xsum} lists the dataset sizes.
\begin{table*}[t!]
    \centering
    \begin{tabularx}{0.8\columnwidth}{|>{\raggedright}X|X|X|}
     \hline
     \textbf{Dataset} & \textbf{CNNDM} & \textbf{XSum}
     \\
     \hline
     \hline
     Train         & 196735 & 186589 \\
     \hline
     Validation    & 13367 & 11316 \\
     \hline
     Test          & 11490 & 11334 \\
     \hline
    \end{tabularx}
    \caption{Preprocessed dataset sizes.
    \label{app:tab:dataset_size_cnndm_xsum}}
\end{table*}

\subsection{Key Implementation}
\label{app:sect:key_implementation}
\subsubsection{Initialization of Extended Modules}
\paragraph{Optimal Transport Plan}
The bilinear transformation weight matrix is initialized to 1.0.
\paragraph{Entity Probability Estimation for AJER}
For named entities in summaries, we directly use the decoder's output logits in formulating AJER. 
For named entities in source documents, we clone the BART-large decoder's logits head (a linear transformation layer) and initialize it using BART-large's weight initialization function.
Note that we do not share the logits linear layer of the decoder for the source document entity probability estimations because we found that sharing is detrimental to summary generation.
\paragraph{Learning Objective Weights}
All learning objective weights $\alpha_{\mathrm{OT}}$, $\alpha_{\mathrm{ANIG}}$, and $\alpha_{\mathrm{JE}}$ are default to 1.0 in our experiments.

\subsubsection{Index Mapping from Named Entity Words to Encoded Tokens}
\label{app:sect:word_level_feature_to_token_mapping}
The word-level named entities and the model-encoded entity tokens are not aligned by their sequential position indices.
Therefore, we map entity words to entity tokens during fine-tuning by utilizing the word-token map and their positions in the document prepared in \Cref{app:sect:data_preparation}.

\subsubsection{Dropout}
Dropout is a widely adopted technique to mitigate overfitting.
The pre-trained BART-large models on CNNDM and XSum from Hugging Face are not configured with activation dropouts.
We thus set the model activation dropouts to 0.1 for fine-tuning the models with our methods, following the dropout configuration of the pre-trained BART-base from Hugging Face.

\subsubsection{Single Token Entity Case for Negative Information Gain}
In such cases, we skip them without incurring costs.

\subsubsection{Early Stop Fine-Tuning Criterion}
Fine-tuning sessions on the BART-large backbones are time-consuming and subject to shared computational resource scarcity. 
However, we would like to take model checkpoints that provide us with good indications of the performance of test inference later, within our computational time budget.
To this end, we use ROUGE metrics to evaluate models on the validation datasets each epoch during fine-tuning, just as we would evaluate models on the test dataset during test inference.
This allows us to take model checkpoints on the highest validation ROUGE scores when the following three or four consecutive validation runs do not have higher ROUGE scores.

\subsubsection{Fine-Tuning on Multi-GPUs}
Our multi-GPU running procedure is based on the reference runtime script\footnote{\label{bart-run-script}\url{https://github.com/huggingface/transformers/blob/v4.9.2/examples/pytorch/summarization/run_summarization_no_trainer.py}.}, including the optimizer settings (the AdamW optimizer, the learning rate $5e^{-5}$ with a linear decay).
The weight decay $1e^{-6}$ is initially adopted from \citet{song_structure-infused_2018} and has worked well in our early projects, which experimented with different weight decay settings.
We set the total mini-batch size to 64.
\par
The models are trained by our methods on two-GPU parallelism. Each GPU card is NVIDIA RTX8000/48GB.
A configuration of OT+AJER has a model size of 1750.418MB.
As models are trained on shared computational resources, the training time may vary considerably due to GPU utilization loads. At a light GPU load, a training iteration may take about one second.

\begin{table*}[t!]
    \centering
    \begin{threeparttable}
    \begin{tabularx}{0.8\columnwidth}{
        |>{\raggedright}p{1.8in}|X|X|
    }
     \hline
     \textbf{Setting} & \textbf{CNNDM} & \textbf{XSum} \\
     \hline
     \hline
     Maximum Length    & 142 & 62 \\
     \hline
     Minimum Length    & 56  & 11 \\
     \hline
     Beam No.          & 4   & 6 \\
     \hline
     Length Penalty    & 2.0 & 1.0 \\
     \hline
    \end{tabularx}
    \caption{Inference settings.}
    \label{tab:inference_settings}
    \end{threeparttable}
\end{table*}
\subsubsection{Inference Setting}
\label{app:sect:key_inference_settings}
We extract the inference settings from Hugging Face's pre-trained BART-base configuration as shown in \autoref{tab:inference_settings}.

\subsection{Human Evaluation of Summary Informativeness}

\subsubsection{Human Evaluation Guidelines}
\label{app:subsect:definition_of_criteria}
We define the following guidelines for the evaluation:
\begin{enumerate}
    \item Evaluation considers the coverage of focal information (explication rather than word-by-word match).
    \item Reference summaries are unnecessarily ideal. Sometimes, they could be uninformative or erroneous. Thus, evaluation should be based on reference summaries consolidated with the key information the articles convey instead of relying on reference summaries alone.
    \item Generated summaries may contain factual errors that are inconsistent with or contradict the facts in articles. Some errors may obscure informativeness more than others. An annotator is free to decide on that.
    \item Generated summaries may contain extrinsic knowledge that is not evidenced in the articles and reference summaries. Such extrinsic knowledge may or may not be factual. An annotator can ignore them.
    \item Information in a model-generated summary may not be evidenced directly in the article sometimes, but it may be deduced, induced, or derivable from the article.
    \item Long summaries are truncated according to modeling configurations. So, the evaluation assesses summaries up to the point.
    \item Summaries may not be in grammatically perfect format, such as whitespace, punctuation, indefinite article, and plurality. They are less concerned if they do not alter the meaning.
    \item If unsure of which model-generated summary is better, please choose 'tie'.
    \item External resources, such as dictionaries, Google Search, and Google Maps, can be used to assist the evaluation.
\end{enumerate}

\begin{figure}[t!]
  \centering
  \includegraphics[width=0.90\columnwidth]{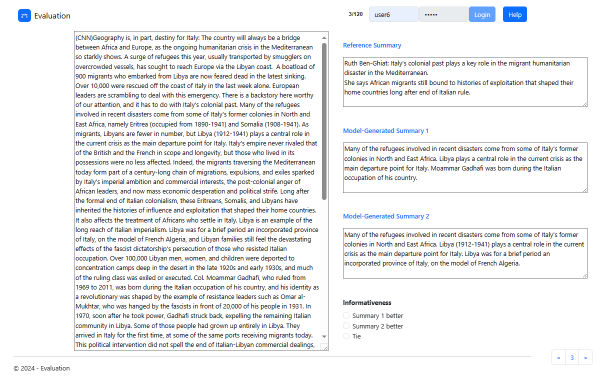}
  \caption{Main user interface of informativeness evaluation.}
  \label{fig:eval_tool_main_page}
\end{figure}

\subsubsection{Normalize Whitespace in Model-Generated Summaries}
\label{post_formatting_summaries_for_informativeness_evaluation}
As discussed in \Cref{app:sect:data_preparation}, our NER-annotated training dataset is built from Penn Treebank segmented words.
We have observed that the Penn Treebank segmented words are compatible well with the pre-trained BART-large model tokenization vocabulary.
However, the model tokenizer has an encoding strategy in which the whitespace preceding a word is converted into a special character and merged with the word before applying byte-pair encoding to generate the token sequences. As a result, the model tokenizer's decoding process in inference restores both whitespace and the word.
This leaves the model trained with the training datasets to insert whitespaces around these tokens, such as around hyphens and inside round brackets, when generating summaries on test sets.
These extra whitespaces do not alter comprehension but might distract an annotator from a fair assessment. Also, to avoid informativeness evaluation discern generated summary-model binding due to such unusual spacing patterns, we therefore `normalize' (remove) these whitespaces in the samples used for evaluation.
Additionally, we have also observed that whitespace after an apostrophe may occasionally be absent in model-generated summaries, including those generated by the baseline.
Similarly, they have little effect on comprehension but could be distractive to an annotator. So, we also `normalize' (insert) whitespace after an apostrophe when it is proper. We do so for the sample summaries of both ours and the baseline.

\subsubsection{Human Evaluation User Interface and Procedure}
\label{app:subsect:human_evaluation_user_interface}
\autoref{fig:eval_tool_main_page} shows the main user interface of the informativeness evaluation tool.
The main page presents an article, a reference summary, and model-generated summaries, followed by the informativeness choices. An annotator can either select the better informative summary or rank the summaries as a `tie'.
60 random samples from the respective CNNDM and XSum test sets are selected.
A total of 120 samples are shown to an annotator by pages.
The tool presents randomly shuffled generated summaries to ensure the evaluation is as fair as possible.
Thus, the model-generated summaries are shown in ``Summary 1" and ``Summary 2" in such a way that the numbers ``1" and ``2" in the captions have no fixed binding to the models throughout the evaluation.
The user interface also does not indicate which dataset a sample is drawn from.
The tool records an annotator's choices in a back-end database from which the queried analysis is conducted upon evaluation completion.

\subsection{Extrinsic but Factual Examples}
\label{app:subsect:extrinsic_factual_examples}
\Cref{app:tab:extrinsic_but_factual_samples_by_bart_model_on_xsum} lists five extrinsic but factual entity-related summaries found in the human evaluation of factuality on XSum.

\newpage
\onecolumn
\setlength{\LTcapwidth}{\linewidth}
\begin{ThreePartTable}
\begin{center}
\small
\def\arraystretch{1.5}
\begin{xltabular}{\textwidth}{|p{3.cm}|X|}
    \hline
    \multicolumn{1}{|c|}{\textbf{Source}} & \multicolumn{1}{c|}{\textbf{Text}}
    \\
    \hline
    \hline
    \endfirsthead
    \hline
    \multicolumn{1}{|c|}{\textbf{Source}} & \multicolumn{1}{c|}{\textbf{Text}} 
    \\
    \hline
    \endhead
    \multicolumn{2}{|r|}{{Continued on next page}}
    \\
    \hline
    \endfoot
    \hline
    \caption{Extrinsic but factual examples (from the 60 model-generated random samples on XSum). 
    We use ellipses to omit the long content that is irrelevant to the discussion.
    The extrinsic but factual entities are highlighted with blue-colored underlines.
    For legibility, we replace the Unicode `\textbackslash{u00a3}' with \pounds, and `\textbackslash{n}' with the Latex's newline format command.
    \label{app:tab:extrinsic_but_factual_samples_by_bart_model_on_xsum}}
    \endlastfoot
    \hline
    Article &
    In only his second season with Porsche, Webber and his two team-mates, German Timo Bernhard and New Zealander Brendon Hartley, need a fourth-place finish or better in the 6 Hours of Bahrain on Saturday to complete a remarkable run of success for car number 17. 
    ... Webber, 39, cannot remember such a hot streak of form beyond his days in junior categories and admits he never expected to be in contention so quickly at this stage of his return to sportscars.\newline "I'm very surprised we're in this position," he said. "We had a few tough races at the start of the championship and now here we are in Bahrain ready to close the championship off hopefully.\newline "It started in an incredible June where we got two cars home at Le Mans and had a 1-2 against Audi.\newline "Since then we have won every race and I don't think we envisaged that - the aggressive mentality of the F1 approach."
    ...
    \\
    \hline
    Reference & 
    Almost exactly five years after Mark Webber lost his best chance of becoming Formula 1 world champion, the Australian is on the brink of fulfilling his title dream in the World Endurance Championship this weekend in Bahrain.
    \\
    \hline
    BART-large & 
    Former Red Bull driver Mark Webber says he is "surprised" his Porsche team are in such a strong position to win the world championship.
    \\
    \hline
    BART-large* /OT+AJER & 
    Former Red Bull driver Mark Webber says he is surprised how good a run of form his Porsche team have been in this year's World Endurance Championship is as he aims to close the gap on championship leader \blueuline{Nico Rosberg}.
    \\
    \hline
    \hline
    Article &
    Levein, the club's director of football, described finishing fifth in the Premiership, as "frustrating and a tad disappointing".\newline "I've got high hopes for him [Cathro]," Levein said. "This is a tough place to manage, as I know myself.\newline "The thing about here is the players like him."\newline Levein points to mistakes in the January transfer window which saw Hearts sign nine players - six of whom have since left the club - as a chief reason for the underwhelming league campaign.
    ... "Rushed is probably a fair assessment," Levein explained.
    ... "For him to be a real success here we need to give him time and let him understand what this is all about," Levein explained. ...
    \\
    \hline
    Reference & 
    Craig Levein retains "high hopes" for Ian Cathro's Hearts tenure but says the club will seek to "repair the damage" of recent transfer windows.
    \\
    \hline
    BART-large & 
    Hearts head coach Ian Cathro has been "rushed" into taking charge of the club, according to director of football Craig Levein.
    \\
    \hline
    BART-large* /OT+AJER & 
    Hearts director of football Craig Levein says head coach Ian Cathro has been "rushed" into taking charge at \blueuline{Tynecastle}.
    \\
    \hline
    \hline
    Article &
    Killie moved eight points clear of bottom side Dundee United with a 3-0 win over St Johnstone on Saturday.
    ..."You want to be part of that and it gives me the chance to attract the kind of players I would like to make sure the club isn't in this situation again.\newline "The club has been dicing with survival for many seasons and that needs to change."
    ...
    \\
    \hline
    Reference & 
    Manager Lee Clark is urging Kilmarnock to build on the first win of his tenure and make sure they are part of an exciting top flight next season.
    \\
    \hline
    BART-large & 
    Hibernian manager Lee Clark hopes his side can regain the "wow factor" of the Scottish Premiership.
    \\
    \hline
    BART-large* /OT+AJER & 
    Manager Lee Clark hopes Kilmarnock can help restore the "wow factor" to the Scottish Premiership as they bid to avoid \blueuline{relegation}.
    \\
    \hline
    \hline
    Article &
    Property developers Michael and John Taggart are offering a settlement where they would repay less than 1p for every pound they owe.\newline That arrangement, known as an  Individual Voluntary Arrangement (IVA), would allow them to avoid bankruptcy.
    ...
    \\
    \hline
    Reference & 
    Two County Londonderry brothers facing bankruptcy owe their creditors up to \pounds213m, the High Court has been told.
    \\
    \hline
    BART-large & 
    Creditors of the Taggart family are set to vote on a proposal that would allow them to avoid bankruptcy.
    \\
    \hline
    BART-large* /OT+AJER & 
    Two of \blueuline{Northern Ireland}'s biggest businessmen are seeking to avoid bankruptcy by offering a voluntary repayment plan to their creditors.
    \\
    \hline
    \hline
    Article &
    Jack Beales, 93, of Rhyl, carried out the attacks while she was a young girl, mainly when he was in his 70s. 
    ...
    The court heard the abuse began when she was aged six and continued for about 10 years.
    ...
    But she said it had not and she was still on strong medication for depression and anxiety, and was awaiting counselling.
    ...
    Det Insp William Jones, of North Wales Police, said: "John (Jack) Beales was a manipulative and depraved sexual predator who committed multiple offences over a sustained period of time. He is now thankfully behind bars.
    \\
    \hline
    Reference & 
    The victim of a pensioner jailed for 13 years for horrific sex attacks has told of the abuse that ruined her life.
    \\
    \hline
    BART-large & 
    A "manipulative and depraved sexual predator" who raped and sexually abused a woman over a 10-year period has been jailed for 12 years.
    \\
    \hline
    BART-large* /OT+AJER & 
    A "depraved" \blueuline{Denbighshire} pensioner who raped and sexually abused his victim as a child has been jailed for 12 years.
    \\
    \hline
\end{xltabular}
\end{center}
\end{ThreePartTable}